\pdfoutput=1

\documentclass[11pt]{article}

\usepackage[final]{acl}

\usepackage{times}
\usepackage{latexsym}
\usepackage{booktabs}

\usepackage[T1]{fontenc}

\usepackage[utf8]{inputenc}

\usepackage{microtype}

\usepackage{inconsolata}

\usepackage{graphicx}

\usepackage{CJKutf8}
\usepackage{makecell}
\usepackage{multirow}
\usepackage{amsmath}
\usepackage{algorithm}
\usepackage{algpseudocode}
\usepackage{float}
\usepackage{colortbl}
\definecolor{myblue}{rgb}{0.53, 0.81, 1.0}

\usepackage{paralist}

\makeatletter
\def\whline#1{%
	\noalign{\ifnum0=`}\fi\hrule \@height #1 \futurelet
	\reserved@a\@xhline}

\title{TongGu: Mastering Classical Chinese Understanding with Knowledge-Grounded Large Language Models}

\author{
 \textbf{Jiahuan Cao\textsuperscript{1,3}}~~
 \textbf{Dezhi Peng\textsuperscript{1}}~~
 \textbf{Peirong Zhang\textsuperscript{1}}~~ 
 \textbf{Yongxin Shi\textsuperscript{1}}~~  
\\
 \textbf{Yang Liu\textsuperscript{1,3}}~~  
 \textbf{Kai Ding\textsuperscript{2,3}}~~  
 \textbf{Lianwen Jin\textsuperscript{1,3}\thanks{Corresponding Author.}}
\\
\\
 \textsuperscript{1}South China University of Technology
\\ \textsuperscript{2}Intsig Information Co., Ltd.
\\ \textsuperscript{3}INTSIG-SCUT Joint Lab on Document Analysis and Recognition
\\
 \small{
   \href{jiahuanc@foxmail.com}{jiahuanc@foxmail.com}~~
   \href{eelwjin@scut.edu.cn}{eelwjin@scut.edu.cn}
 }
}

\begin{document}
\maketitle
\begin{abstract}

Classical Chinese is a gateway to the rich heritage and wisdom of ancient China, yet its complexities pose formidable comprehension barriers for most modern people without specialized knowledge. 
While Large Language Models (LLMs) have shown remarkable capabilities in Natural Language Processing (NLP), they struggle with Classical Chinese Understanding (CCU), especially in data-demanding and knowledge-intensive tasks. In response to this dilemma, we propose \textbf{TongGu} (mean understanding ancient and modern), the first CCU-specific LLM, underpinned by three core contributions. First, we construct a two-stage instruction-tuning dataset ACCN-INS derived from rich classical Chinese corpora, aiming to unlock the full CCU potential of LLMs. Second, we propose Redundancy-Aware Tuning (RAT) to prevent catastrophic forgetting, enabling TongGu to acquire new capabilities while preserving its foundational knowledge. Third, we present a CCU Retrieval-Augmented Generation (CCU-RAG) technique to reduce hallucinations based on knowledge-grounding. Extensive experiments across 24 diverse CCU tasks validate TongGu's superior ability, underscoring the effectiveness of RAT and CCU-RAG. 
The model and dataset are available at \url{https://github.com/SCUT-DLVCLab/TongGu-LLM}.

\end{abstract}

\section{Introduction}

Classical Chinese is a vital bridge in connecting the present with the wisdom of ancient China, illuminating insights into historical social life and cultural practices. However, the significant linguistic differences between classical and modern Chinese, including vocabulary and syntax, render this invaluable heritage prohibitively challenging for non-experts to understand.


Recent advancements in Large Language Models (LLMs) have demonstrated remarkable capability in addressing various Natural Language Processing (NLP) tasks \cite{raffel2020T5, zhang2022opt, chung2022flan-t5, chowdhery2023palm, brown2020gpt3, touvron2023llama, touvron2023llama2, openai2023gpt4}, prompting researchers to explore their capabilities in the specialized realm of Classical Chinese Understanding (CCU). 
However, existing models, including general-purpose and preliminary CCU-specific LLMs \cite{chunhua_hf,xunzi_github}, often struggle with tasks that require large-scale training data or extensive domain knowledge.
This predicament primarily stems from two causes: the lack of dedicated instruction-tuning datasets capable of unleashing their full capabilities, and models' innate  propensity to generate hallucinations when tackling knowledge-intensive tasks without sufficient factual grounding.

To address these challenges, we present \textbf{TongGu}, a pioneering vertical domain LLM as well as the most proficient CCU specialist. 
We commence with devising an automated pipeline to construct instruction data from classical Chinese texts, resulting in \textbf{ACCN-INS} (short for ancient Chinese instruction), the first publicly available CCU instruction dataset catering to diverse CCU tasks.
Subsequently, TongGu undergoes a two-stage instruction tuning, respectively for the optimization of data-hungry and data-efficient tasks. 
Here, "data-hungry" means that a substantial volume of data is required to attain satisfactory model performance, and "data-efficient" denotes that a small amount of data is sufficient to achieve desired outcomes.
It is first fine-tuned on data-hungry tasks using large-scale training data, such as classical to modern Chinese translation, followed by a second stage of fine-tuning on data-efficient tasks such as punctuation with small-scale data.
To prevent catastrophic forgetting during the two-stage fine-tuning, we propose \textbf{Redundancy-Aware Tuning (RAT)}, a novel sparse fine-tuning (\textit{a.k.a.} Parameter-Efficient Fine-Tuning (PEFT)) method that identifies and freezes the most crucial layers for the current task according to layer redundancy. 
RAT effectively injects new capability to the model while preserving prior learnt knowledge, thereby ensuring the stability and retention of foundational knowledge in TongGu. 
In addition, we propose an efficient \textbf{CCU Retrieval-Augmented Generation (CCU-RAG)} method that significantly mitigates the propensity of hallucinations in knowledge-intensive tasks, further bolstering TongGu's performance.

To summarize, our contributions are as follows:

\begin{itemize}
    \item We develop TongGu, a pioneering vertical domain LLM adept at managing a broad spectrum of CCU tasks.
    \item We design a pipeline for automatically generating instruction data from classical Chinese texts and construct the ACCN-INS dataset, the first classical Chinese instruction data publicly available.
    \item We propose Redundancy-Aware Tuning (RAT), a sparse fine-tuning method to alleviate catastrophic forgetting in the two-stage fine-tuning.
    \item To reduce hallucinations in knowledge-intensive classical Chinese tasks for LLMs, we introduce a task-specific efficient Retrieval-Augmented Generation (RAG) method.
\end{itemize}

\section{Related Work}
\subsection{Large Language Models}

Large Language Models (LLMs), such as GPT-4 \cite{openai2023gpt4}, LLaMA \cite{touvron2023llama}, and Baichuan \cite{baichuan2023baichuan2}, have exhibited unprecedented prowess across numerous NLP tasks. Thanks to dedicated training techniques like instruction tuning, current LLMs can develop not only exceptional general intelligence but also commendable domain-specific specialties. In this context, research into vertical domain LLMs \cite{roziere2023codellama, wu2023pmcllama, yunxiang2023chatdoctor} emerges as a burgeoning topic and continually fuels endeavors within the community.

\subsection{Language Modeling for Classical Chinese Understanding}
Early Classical Chinese Understanding (CCU) systems were trained for specific tasks such as translation \cite{jiang2023C2C, chang2021time-aware}, punctuation \cite{li2009punctuation}, and named entity recognition (NER) \cite{yu2020bert-ner, han2018chinese-term}. 
However, these methods often relied heavily on large amounts of manually annotated data to achieve decent performance. 
GujiBERT \cite{wang2023gujibert} utilized large-scale unlabeled classical Chinese corpora for masked pre-training, providing task-specific models with embeddings that encode classical Chinese knowledge. 
SikuGPT \cite{chang2023sikugpt} leveraged massive classical Chinese corpora for generative pre-training, highlighting the potential of generative pre-training in classical Chinese poetry and prose creation. 
Bloom-7b-Chunhua \cite{chunhua_hf} and Xunzi-Qwen-7B-CHAT \cite{xunzi_github}, which combined an open-source base model with large-scale classical Chinese corpora, preliminarily investigated the classical Chinese language understanding capabilities of LLMs.


\begin{figure*}
\centering
\includegraphics[scale=0.3]{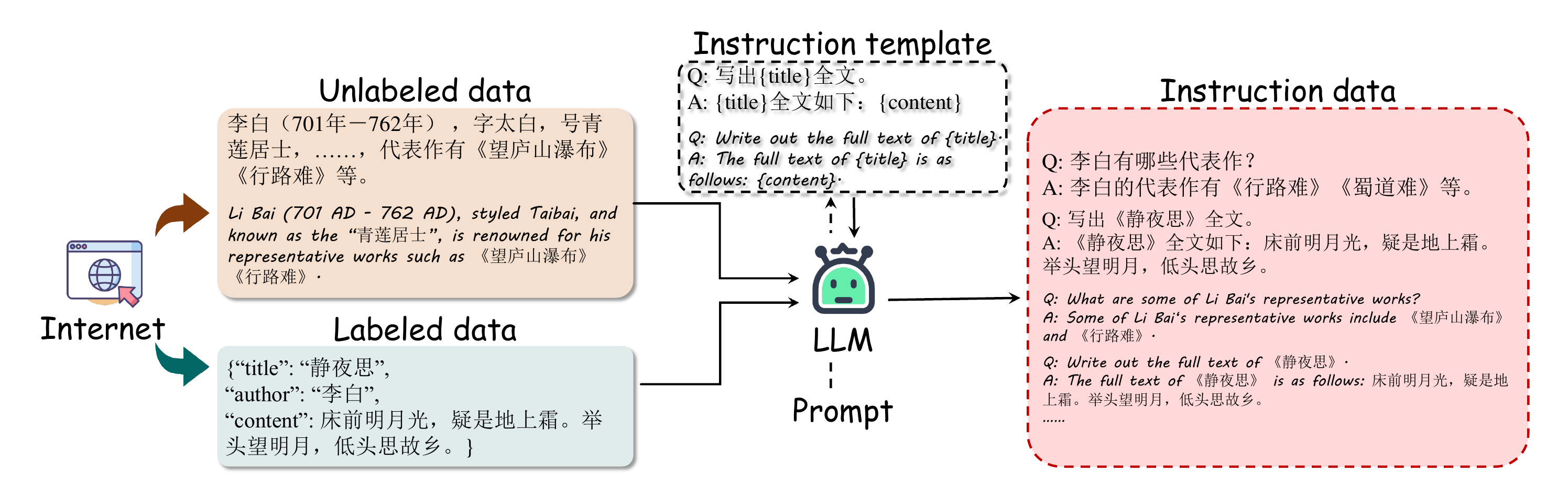}
\caption{Overview of the CCU instruction data generation pipeline from labeled and unlabeled text.}
\label{fig:ACCN-INS}
\end{figure*}

\subsection{Retrieval-Augmented Generation}

Retrieval-Augmented Generation (RAG) is a technique that enhances generation models by incorporating relevant content retrieved from knowledge sources, which has been proven to be effective for mitigating hallucinations in LLMs \cite{guu2020realm, lewis2020rag}.

RAG system generally adheres to a workflow encompassing three components: indexing, retrieval, and generation. 
Recent advancements have focused on enhancing the retrieval component.
RETRO \cite{borgeaud2022retro} amalgamates large-scale corpora with pre-trained frozen BERT embeddings. 
Atlas \cite{izacard2022atlas} conducts joint training of a retriever and a sequence-to-sequence model to attain a language model with robust few-shot learning capabilities.
Self-RAG \cite{asai2024selfrag} selectively retrieves knowledge and generates critique tokens to criticize its own output.

Apart from prior works, our proposed CCU-RAG specifically targets knowledge-intensive tasks within classical Chinese question-answering. It centers on endowing TongGu to discern both the timing and content of retrieval, while enhancing the synergistic efficiency between the model and the retrieval system it relies upon.

\section{ACCN-INS Dataset}

The complexity of manual annotation in classical Chinese Question Answering (QA) tasks requires extensive human expertise, resulting in labor-intensive processes.
To alleviate the labor intensity, harnessing LLM for automatic data annotation becomes a natural and efficient solution. 
Nevertheless, LLM harbors the propensity to inadvertently introduce inaccuracies during the data generation process. 
To address this issue, we present a semi-automated annotation method that combines classical Chinese corpora using aligned LLMs, thereby producing reliable instructional data for specialized CCU tasks.
 
Firstly, we collected classical Chinese corpora from multiple public sources, such as Daizhige \cite{daizhige_github}, textbooks, and examination papers. 
We then design a pipeline to transform these data into the instructional format, as depicted in Figure \ref{fig:ACCN-INS}.

\begin{table}[H]
\centering
\caption{Statistics of the generated data.}
\resizebox{0.8\linewidth}{!}{
\begin{tabular}{lc}
\whline{1.1pt}
\textbf{Statistics} \\
\hline
\# instructions & 4,020,136 \\
\quad\# instructions from labeled data & 4,014,355 \\
\quad\quad\# data-hungry tasks data & 4,000,000 \\
\quad\quad\# data-efficient tasks data & 14,355 \\
\quad\# instructions from unlabeled data & 5,781 \\
\quad\quad\# data-efficient tasks data & 5,781 \\
\hline
avg. instruction length & 48.59 \\
avg. output length & 68.96 \\
\whline{1.1pt}
\end{tabular}
}
\label{tab:data_statistics}
\end{table}

\textbf{Labeled data.}
Labeled data refers to the data equipped with well-curated labels, such as the dynasty and author of a poem.
Owing to its highly structured organization, a large amount of instruction data can be simply synthesized using instruction templates.
Specifically, for each task, we first provide 8 instruction examples handcrafted by human experts as in-context examples, prompting the aligned LLM to generate a broader range of diverse instruction templates. 
Finally, we can populate the structured data into these instruction templates to obtain instruction data in the QA format.

\textbf{Unlabeled data.}
Unlabeled data refers to unlabeled text segments, such as introductions to certain poets, where information like era, life experiences, and representative works are intermingled within the same text segment. 
We adopt a reading comprehension approach, treating the unlabeled text segments as reference materials and requiring the aligned LLM to extract QA pairs from them. Similarly, we use 8 human-written QA pairs as in-context examples.

\begin{figure*}[ht]
\centering
\includegraphics[scale=0.18]{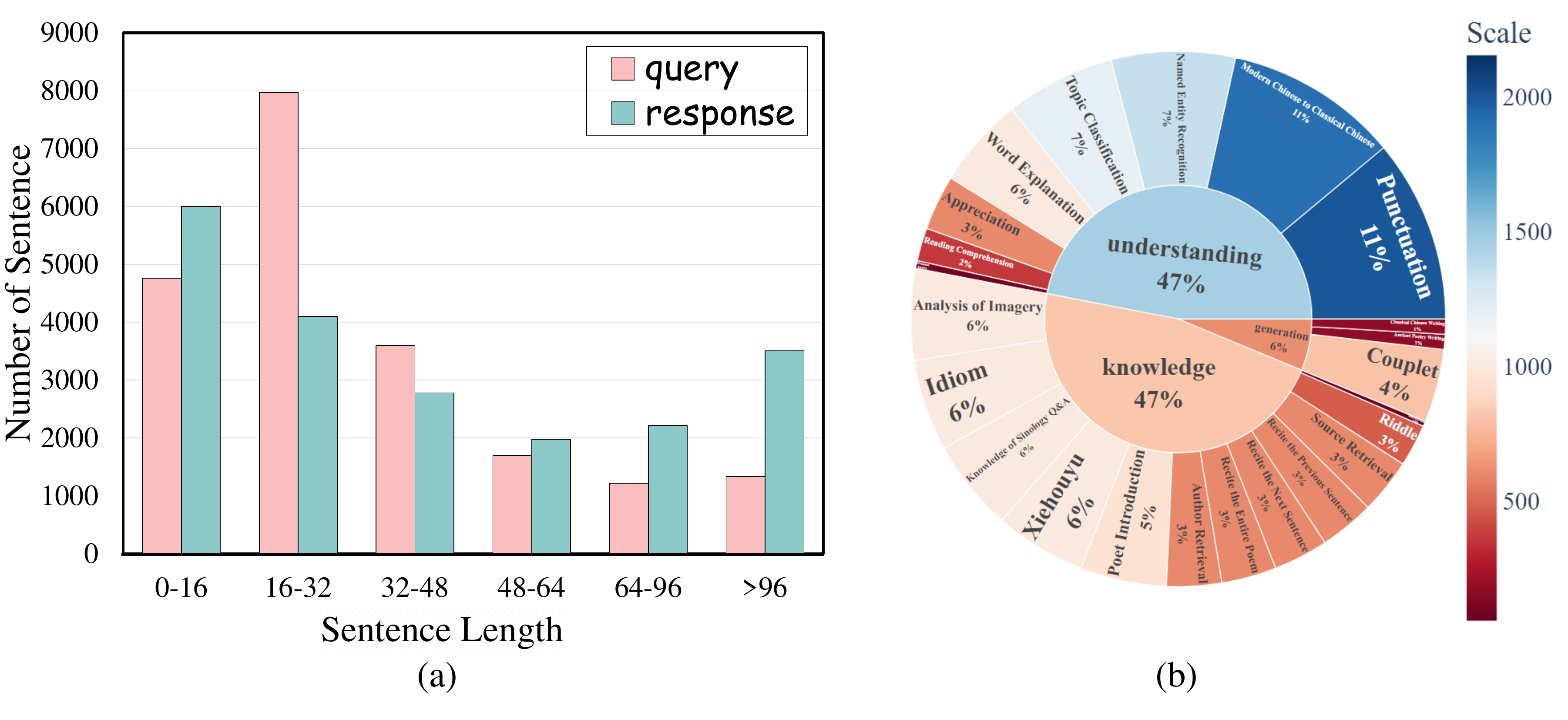}
\caption{Data statistics of data-efficient tasks data in ACCN-INS dataset. (a) Distribution of sentence length. (b) Sample distribution for each task. Zoom in for a better view.}
\label{fig:dis_length_domain}
\end{figure*}


Employing the proposed data generation pipeline and leveraging ChatGPT \cite{ouyang2022instruct-GPT} as the aligned model, we initially generated a large corpus of classical Chinese instruction data. To ensure the quality and accuracy of this generated data, we implemented a rigorous cleaning process. This process included several key steps: removing duplicate entries, standardizing punctuation marks to ensure consistency between Chinese and English usage, and carefully reviewing the content for accuracy and appropriateness. After this meticulous verification and refinement process, we obtained a final dataset of 4,020,136 instances of high-quality classical Chinese instruction data, among which 4,014,355 instances originated from structured text and 5,781 from unstructured text.
In Table \ref{tab:data_statistics}, we delineate the quantities of data obtained through various generation methods, coupled with the average lengths of instructions and outputs across the entire instruction dataset. 
Data-hungry tasks data in ACCN-INS contains 4,000,000 samples of classical-to-modern Chinese translation corpus, and Figure \ref{fig:dis_length_domain} presents the data statistics of various data-efficient tasks data in ACCN-INS. As illustrated in Figure \ref{fig:dis_length_domain} (a), the length distribution indicates that responses are generally longer than queries, with a multitude of responses exceeding 96 characters. This suggests the rich and comprehensive nature of the information in the ACCN-INS dataset, benefiting the model to develop deeper CCU proficiencies.
The detailed task types and corresponding sample count of ACCN-INS are illustrated in Figure \ref{fig:dis_length_domain} (b), demonstrating the diversity and comprehensiveness of this dataset. 
Detailed examples of each task are included in Appendix \ref{sec:appendix_data}.

\begin{figure*}[ht]
\centering
\includegraphics[scale=0.45]{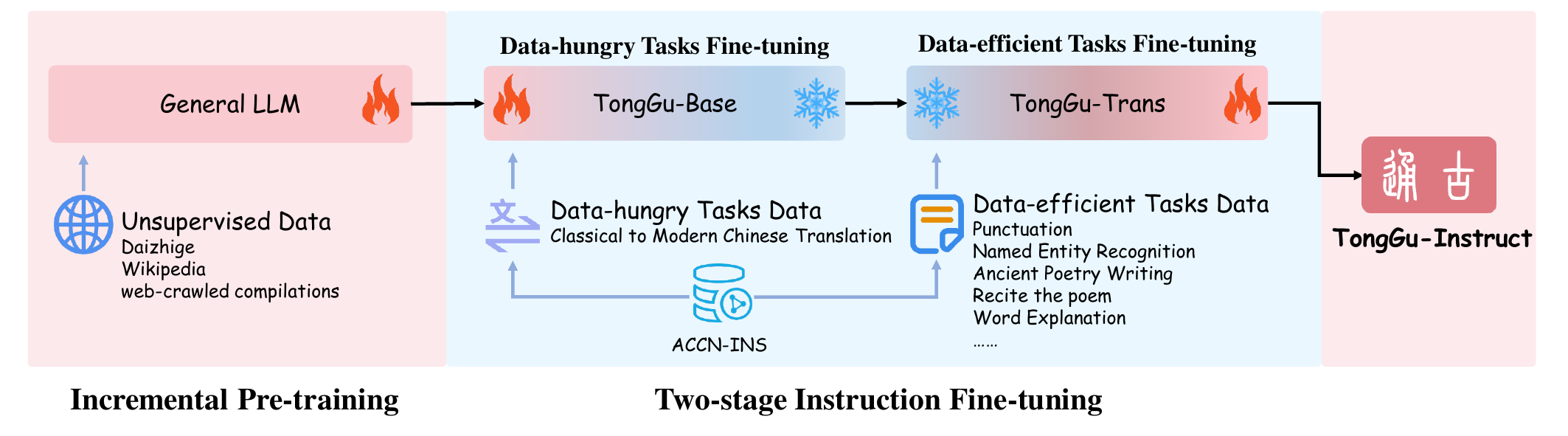}
\caption{Overview of the training pipeline.}
\label{fig:overview_training}
\end{figure*}

\section{TongGu}
TongGu is a generalist LLM specifically designed for Classical Chinese Understanding (CCU), 
whose capability is built based on three core steps as shown in Figure \ref{fig:overview_training}.
First, we perform incremental pre-training on TongGu on a mixed corpus of 4.6 billion tokens consisting of classical and modern Chinese to enrich its CCU knowledge.
Second, TongGu is fine-tuned on millions of instruction data with the proposed PEFT method Redundancy-Aware Tuning (RAT), which not only enhances multi-task understanding proficiency but also ensures highly efficient instruction tuning.
Thirdly, we introduce CCU-RAG, a task-specific Retrieval-augmented generation (RAG) mechanism to alleviate hallucinations in knowledge-intensive tasks.
Through these three steps, TongGu demonstrates its ability to handle 24 diverse CCU tasks effectively, making it a powerful tool for understanding classical Chinese texts.

\begin{table}[ht]
\renewcommand{\arraystretch}{1.2}
\centering
\caption{Incremental pre-training data. For each subset, we list the sampling proportion, disk size, and number of epochs during the training process.}
\resizebox{\linewidth}{!}
{
\begin{tabular}{lcccc}
\whline{1.1pt}
\textbf{Dataset} & \textbf{Sampling prop.} & \textbf{Disk size (GB)} & \textbf{Tokens (B)} & \textbf{Epochs} \\
\hline
\rowcolor{gray!15}
\textit{Classical Chinese} & & & & \\
\hline
Series & 3.16$\%$ & 0.27 & 0.078 & 1 \\
Buddhism & 4.92$\%$ & 0.42 & 0.129 & 2 \\
Confucianism & 4.22$\%$ & 0.36 & 0.112 & 2 \\
Medicine & 1.52$\%$ & 0.13 & 0.042 & 1 \\
History & 17.56$\%$ & 1.5 & 0.453 & 2 \\
Philosophy & 1.17$\%$ & 0.1 & 0.032 & 1 \\
Changes & 0.94$\%$ & 0.08 & 0.026 & 1 \\
Poetry & 8.20$\%$ & 0.7 & 0.222 & 2 \\
Literature & 12.88$\%$ & 1.1 & 0.315 & 2 \\
Taoism & 0.94$\%$ & 0.08 & 0.026 & 1 \\
Art & 4.68$\%$ & 0.4 & 0.014 & 1 \\
\hline
\rowcolor{gray!15}
\textit{Modern Chinese} & & & & \\
\hline
wiki-zh & 39.9$\%$ & 3.4 & 0.968 & 2 \\
\hline
\rowcolor{myblue!15}
\textbf{Total} & - & 8.54 & 2.41 & - \\
\whline{1.1pt}
\end{tabular}
}
\label{tab:pretraining_data}
\end{table}

\subsection{Incremental Pre-training}
For the incremental pre-training of TongGu, we curate hybrid incremental pre-training data consisting of classical Chinese and modern Chinese texts with 2.41 billion tokens in total (with tokenizer from Baichuan2-7B-Base).
Table \ref{tab:pretraining_data} illustrates the various data sources and their respective size and sampling proportions.
\textbf{Classical Chinese:} The classical Chinese text is primarily sourced from Daizhige \cite{daizhige_github} and web-crawled compilations, spanning diverse domains including history, poetry, medicine, and Buddhist studies, \textit{etc}. We employ a data cleaning pipeline inspired by RedPajama \cite{together2023redpajama} to perform text format standardization and document-level deduplication.
\textbf{Modern Chinese:} We utilize the wiki-zh corpus curated by MNBVC \cite{mnbvc} and further perform line-level deduplication.

We utilize the Baichuan2-7B-Base \cite{baichuan2023baichuan2} as a foundational model and performed incremental pre-training based on the curated mixtured data. 
Following the standard language modeling paradigm outlined in GPT \cite{radford2018gpt-1}, we train the model to predict the next token based on the context provided by the previous tokens. As a result, we develop a classical Chinese base model, \textit{TongGu-7B-Base}, poised to serve as a potent foundation for subsequent fine-tuning.
More training details such as information on training duration and hardware specifications are provided in Appendix \ref{sec:appendix_training}, Table \ref{tab:para_incremental_it}. 

\subsection{Two-stage Instruction Fine-tuning}

Different CCU tasks can be categorized as data-hungry and data-efficient based on their data requirements. The former demands an insatiable feast of data to attain satisfactory performance, epitomized by the Translation task between classical and modern Chinese. The latter, however, is capable of achieving satisfactory performance with modest data provisions, such as punctuation restoration or topic classification.
To address both data-hungry and data-efficient task requirements, we conduct a two-stage fine-tuning procedure that first fine-tune TongGu on the data-hungry translation task with a large amount of data and then fine-tune it on data-efficient tasks such as punctuation and topic classification with a smaller scale of data.
Through the progressive fine-tuning, the model can effectively capitalize on large-scale data for the primary translation task, while enabling efficient transfer learning and specialization for multiple tasks with limited data, thus fostering comprehensive CCU task proficiency.

Despite the advantage in terms of general ability fostering of the two-stage fine-tuning, this scheme potentially confronts with the catastrophic forgetting issue. 
To mitigate this issue, we propose a novel PEFT method termed Redundancy-Aware Tuning (RAT). Recent study \cite{gromov2024prune1} have revealed that certain layers in LLMs are highly redundant, suggesting they can be removed without significantly impacting the performance of downstream tasks. 
Building upon this inspiration, RAT identifies and preserves these redundant layers while freezing the others during the training of new tasks. By selectively updating only the redundant layers deemed non-essential for the erstwhile tasks, this approach effectively retain the acquired knowledge, which, therefore, mitigates catestrophic forgetting, while enabling efficient adaptation to new tasks.

Algorithm \ref{alg:rat} summarizes the procedure of RAT.
Initially, we randomly select a portion of training data as a calibration set to extract and monitor the model's internal dynamics. Subsequently, we collect the hidden state representations from each model layer during inference, and calculate the cosine similarity between I/O hidden states in tandem. The cosine similarity between I/O hidden states for the $i^{th}$ layer is calculated as:

\begin{equation}\label{equ:cos}
    \text{CoS}_i = \frac{1}{L} \sum_{t=1}^{L} \frac{H_{i,t} \cdot H_{i+1,t}}{\|H_{i,t}\|_2 \|H_{i+1,t}\|_2}
\end{equation}

\noindent where $H_{i,t}$ represents the hidden state vector at timestep $t$ for layer $i$, $\|\cdot\|_2$ denotes the $L_2$ normalization and \textit{L} represents the sample sentence length. We calculate the mean of the cosine similarity on all samples in the calibration set.

Finally, we freeze the layers exhibiting lower similarity scores. 
Due to the tendency of deeper layers to harbor a greater degree of redundancy, we have implemented a grouping and ranking strategy to avoid potential impairment to the model's learning capacity deriving from solely fine-tuning the deeper layers.
The layers of TongGu are partitioned into \textit{N} groups according to their depth, from the shallowest to the deepest. Within each group, we selectively subject the layer exhibiting the highest redundancy to fine-tuning, while the remaining layers are kept frozen.

\begin{algorithm}[t]
\caption{Redundancy-Aware Tuning (RAT)}
\label{alg:rat}
\begin{algorithmic}[1]
\Require Model $\mathcal{M}$, Calibration Data $\mathcal{D}_{old}$, Training Data $\mathcal{D}_{new}$, Groups $N$
\For{$s \in \mathcal{D}_{old}$}
    \For{$i = 1$ to $L-1$}
        \State $H_{i,t}, H_{i+1,t} \gets \mathcal{M}(s)$
        \State $\text{CoS}_i \gets \frac{1}{L} \sum_{t=1}^{L} \frac{H_{i,t} \cdot H_{i+1,t}}{\|H_{i,t}\|_2 \|H_{i+1,t}\|_2}$
    \EndFor
\EndFor
\State $\text{redundancy} \gets \overline{\text{CoS}_i}$
\State Divide layers into $N$ groups: $\{G_1, \ldots, G_N\}$
\For{$g = 1$ to $N$}
    \State $l_g \gets \arg\max_{l \in G_g} (\text{redundancy})$
    \State Fine-tune $l_g$ on $\mathcal{D}_{new}$, keep others frozen
\EndFor
\end{algorithmic}
\end{algorithm}

\textbf{Data-hungry Tasks Fine-tuning:}
We use the data-hungry tasks' data from ACCN-INS for fine-tuning, resulting in the model named \textit{TongGu-7B-trans}.
\textbf{Data-efficinet Tasks Fine-tuning:}
We proceed to use the data-efficient tasks' data from ACCN-INS to fine-tune the \textit{TongGu-7b-trans} model, cultivating the model's capabilities into a broader range of CCU tasks. Moreover, we filter 10,000 dialogue samples between humans and AI assistants from ShareGPT \cite{sharegpt_hf} as supplementary data, further enhancing the model's conversational abilities.
As a result, we obtained the final model named \textit{TongGu-7B-Instruct}.
To address the issue of catastrophic forgetting during the training process, we employ the proposed RAT method with \textit{N} set to 8 throughout both stages. In the first stage, a subset of the incremental pre-training data serves as the calibration set. In the second stage, a subset of classical-to-modern Chinese translation corpus is used for the same purpose.
More details can be found in Appendix \ref{sec:appendix_training}.

\subsection{CCU-RAG}

In knowledge-intensive CCU tasks, general-purpose LLMs and initial efforts in this field usually suffer from severe hallucinations. 
Recently, Retrieval-Augmented Generation (RAG) has been proven to be an effective solution in mitigating these hallucinations in LLMs \cite{gao2023rag-servey1, zhao2024rag-servey2}. Hence, we propose the CCU-RAG, a task-specific efficient RAG framework to enhance the veracity and reliability of the generated outputs from TongGu.

\begin{figure}[ht]
\centering
\includegraphics[scale=0.35]{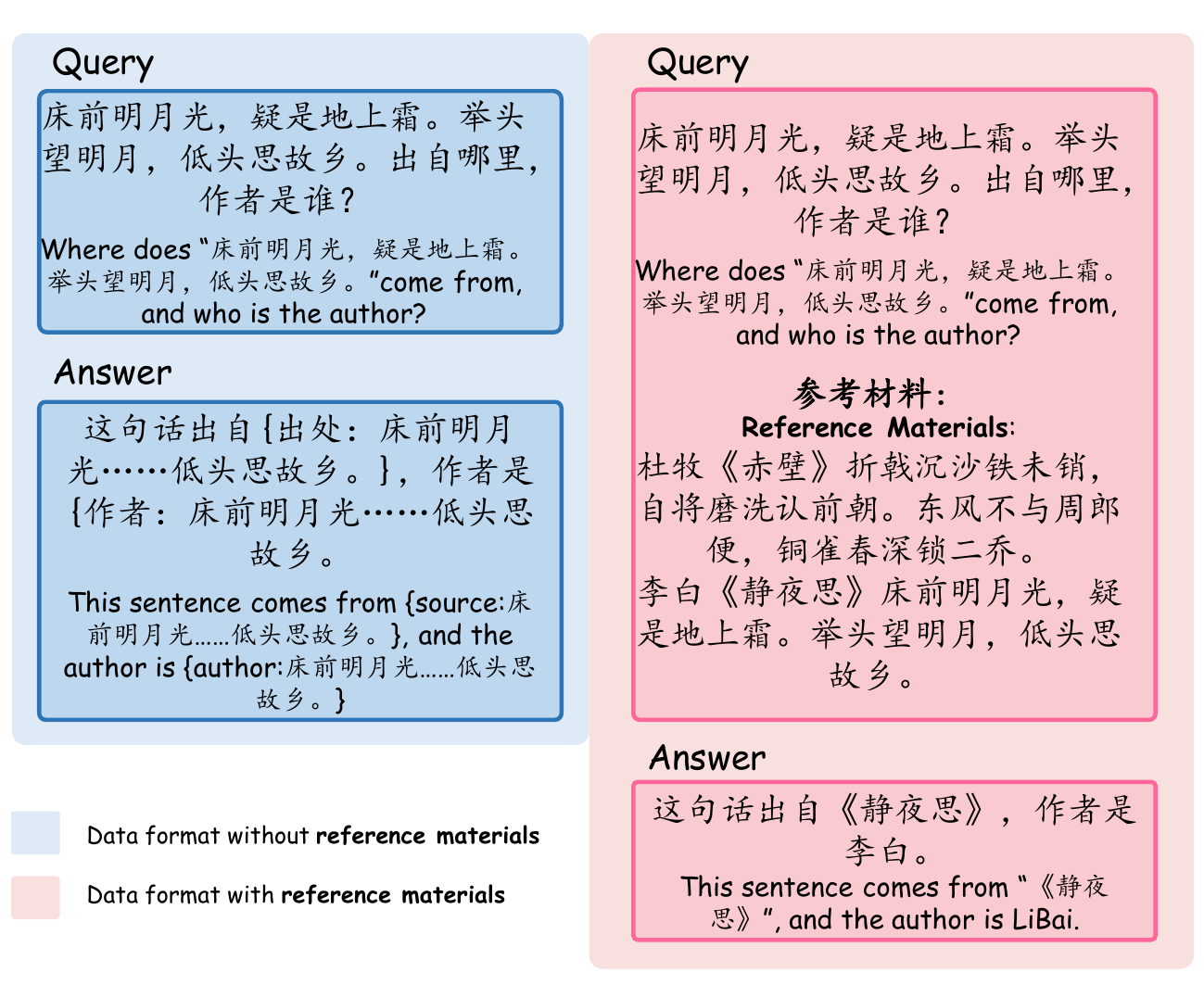}
\caption{Two examples of reformatted knowledge-intensive tasks, with the difference being whether reference materials are provided.}
\label{fig:rag_sample}
\end{figure}

\begin{figure}[ht]
\centering
\includegraphics[scale=0.6]{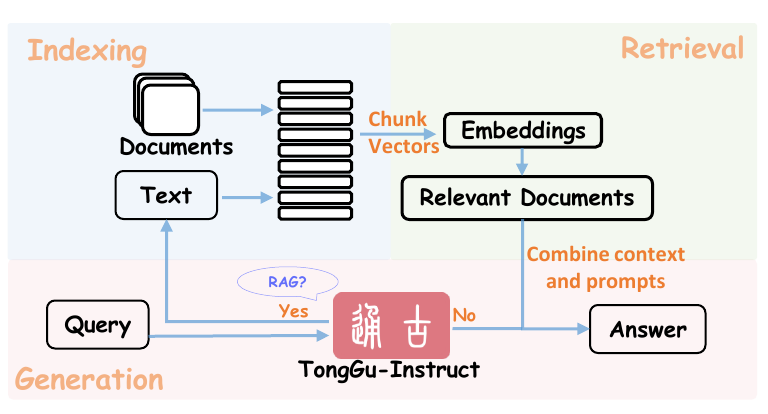}
\caption{Workflow of TongGu response with CCU-RAG.}
\label{fig:rag}
\end{figure}

Firstly, knowledge-intensive samples were extracted from the instruction data, encompassing \textbf{source retrieval, author retrieval, previous sentence recitation, next sentence recitation, entire poem recitation}. 
Subsequently, these samples are reformatted into two types of data to simulate the two steps in RAG, as exemplified in Figure \ref{fig:rag_sample}.
One format retains the original query, with the response being reformulated as multi-level key-value pairs that support searching and retrieving. 
The other format involves appending \textit{reference materials} to the original query, while maintaining the original response. 
The workflow of TongGu with the CCU-RAG system is illustrated in Figure \ref{fig:rag}.
When TongGu receives a user query, if it is a knowledge-intensive task and lacks sufficient relevant knowledge, it generates a multi-level key-value pair to call the retrieval module. Then, the retrieved content is concatenated into the second instruction format and re-entered into the TongGu, enabling it to output more accurate answers. This judgment process is completed by the TongGu itself.
Ultimately, these reformatted samples were utilized to replace the original samples in ACCN-INS, resulting in an enhanced retrieval-augmented instruction fine-tuning database.

\begin{table*}[ht]
\centering
\caption{Evaluation results for various LLMs based on performance on C$^{3}$bench. $^{\star}$ represents our reproduction. \textbf{Bold} indicates the best score, \underline{underline} indicates the second best result.}
\resizebox{\linewidth}{!}{
\begin{tabular}{lcccccc}
\whline{1.1pt}
 \textbf{Model} & \textbf{Classifications} ↑ & \textbf{Retrieval} ↑ & \textbf{NER} ↑ & \textbf{Punctuation} ↑ & \textbf{Translation} ↑ & \textbf{Avg.} ↑ \\
\hline
Bloom-7B-Chunhua & 39.62 & 13.36 & 34.70 & 62.19 & 11.27 & 33.23 \\
Baichuan2-7B-Chat & 37.00 & 18.36 & 63.25 & 53.96 & 13.70 & 37.15 \\
Baichuan2-13B-Chat & 44.26 & 17.79 & 46.67 & 65.11 & 12.45 & 37.26 \\
ChatGLM2-6B & 50.28 & 9.03 & 28.56 & 28.48 & 6.76 & 24.62 \\
Qwen-7B-Chat & 49.65 & 13.92 & 28.33 & 69.61 & 15.61 & 35.42 \\
Qwen-14B-Chat & 44.93 & \underline{25.90} & \underline{66.72} & 71.83 & 15.38 & 44.95 \\
LLaMA2-Chinese-7B-Chat & 18.78 & 3.20 & 12.62 & 34.73 & 4.24 & 14.71 \\
LLaMA2-Chinese-13B-Chat & 28.75 & 2.27 & 9.31 & 47.27 & 5.91 & 18.70 \\
Moss-moon-003-SFT & 15.07 & 15.84 & 28.90 & 58.39 & 13.35 & 26.30 \\
GPT-3.5-turbo & 50.65 & 7.36 & 63.83 & 61.34 & 11.94 & 39.02 \\
GPT-4 & 53.88 & 13.71 & 63.87 & 67.31 & 12.09 & 42.17 \\
ERNIE-bot-turbo & 50.70 & 21.22 & 9.61 & 65.29 & 10.66 & 31.50 \\
Spark-v3 & 51.61 & 21.83 & 53.81 & \underline{85.38} & \underline{34.58} & \underline{49.44} \\
abab5-chat & 52.20 & 15.53 & 34.64 & 65.42 & 10.56 & 35.67 \\
ChatGLM$\_$Turbo & 56.15 & 20.49 & 30.04 & 69.72 & 10.91 & 37.46 \\
Xunzi-Qwen-Chat$^{\star}$ & 42.48 & 11.58 & 54.64 & 78.62 & 16.64 & 40.79 \\
GLM4-9B-Chat$^{\star}$ & 52.99 & 19.73 & 48.15 & 70.96 & 17.26 & 41.82 \\
Qwen2-7B-Instruct$^{\star}$ & \underline{56.22} & 20.60 & 64.36 & 73.19 & 15.17 & 45.91 \\
\hline
\textbf{TongGu-7B-Instruct (Ours)} & \textbf{72.47} & \textbf{77.30} & \textbf{73.46} & \textbf{89.97} & \textbf{54.43} & \textbf{74.53} \\
\whline{1.1pt}
\end{tabular}}
\label{tab:res_models}
\end{table*}

It is worth noting that generating complete key-value pairs for lengthy sentences can be time-consuming. Therefore, we fine-tuned the model to focus solely on generating the beginning and ending fragments of the key-value pairs, using ellipses to replace excessively long intermediate text segments. The complete text in the context is used for retrieval based on the uncompleted text segment generated by the model.  
This approach is simple yet effective, significantly reducing the time required from user input to model response.

\section{Experiments}
In our experiments, we focus on evaluating TongGu's capabilities in Classical Chinese Understanding (CCU) across three dimensions: understanding, generation, and knowledge.

\subsection{Performance on C\texorpdfstring{$^{3}$}{3}bench}
To evaluate the performance of TongGu on common classical Chinese tasks, we utilize the C$^{3}$bench \cite{cao2024c3bench}, a comprehensive classical Chinese benchmark designed for LLMs, covering ten domains and five common classical Chinese tasks. 

\begin{figure}[t]
\centering
\includegraphics[scale=0.3]{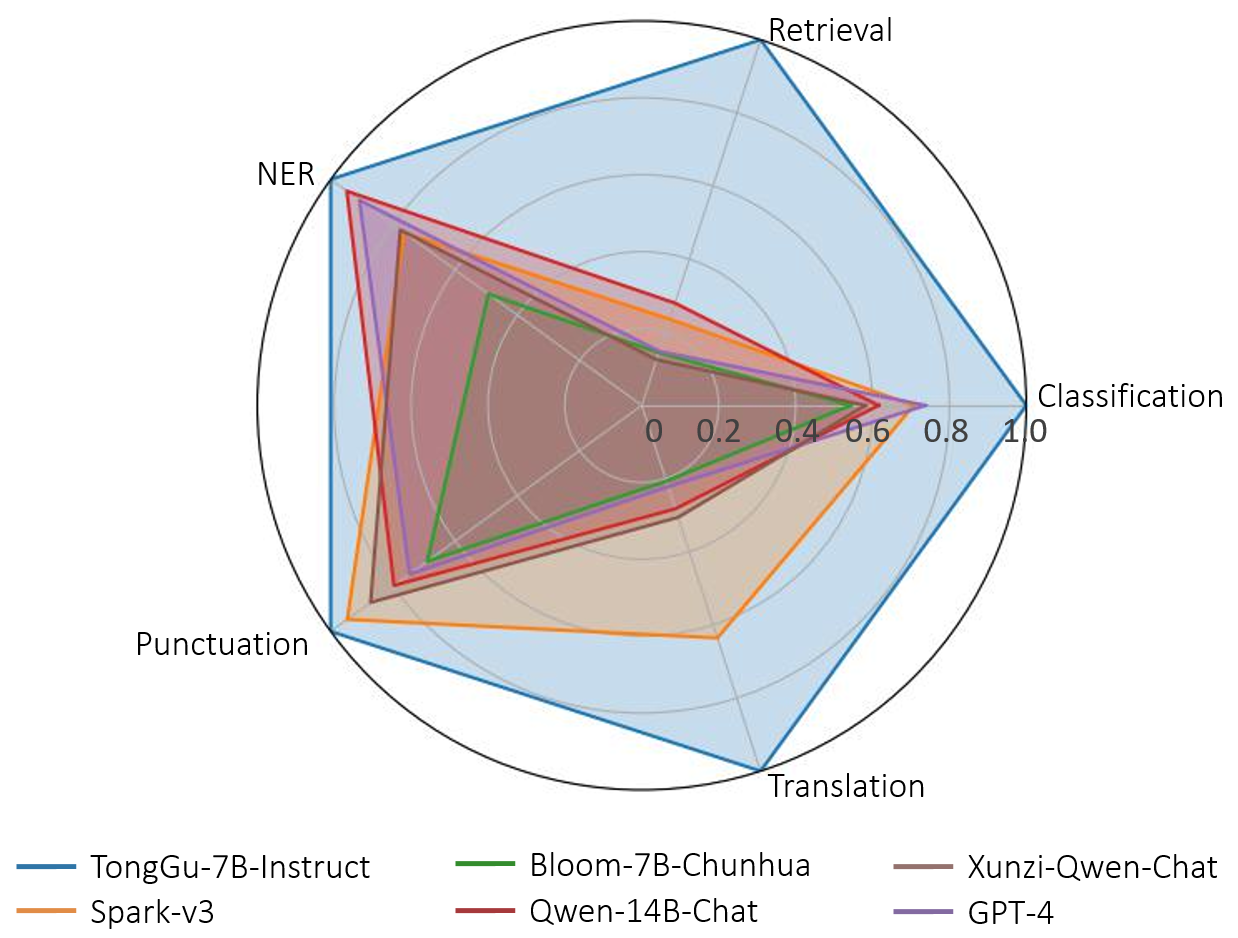}
\caption{Radar charts of performances on C$^{3}$bench. The values have been normalized to a 0-1 scale using the metrics of TongGu-7B-Instruct.}
\label{fig:leaderboard}
\end{figure}

The details of C$^{3}$bench is presented in Appendix \ref{sec:appendix_evaluation}, Table \ref{tab:c3_details}.
We rigorously adhere to the settings outlined in the C$^{3}$bench paper and conduct a zero-shot evaluation. 
The quantitative results of TongGu's performance on the C$^{3}$bench are presented in Table \ref{tab:res_models}, alongside the more radar charts in Figure \ref{fig:leaderboard}. 
As confirmed by the outcomes, we can observe that TongGu outperforms existing LLMs across all five tasks, especially in knowledge-intensive retrieval task and data-hungry translation task.

\subsection{Performance on Broader CCU Tasks}
The ACCN-INS dataset covers a broader range of tasks than existing benchmarks, including tasks such as poetry creation and Flying flower order (a form of Chinese literary game requiring a poem containing a certain keyword). 
For a comprehensive evaluation, we generate a test set of 1,600 samples using the same process as the training set, utilizing it only during model evaluation. 
The details of our test dataset are presented in Appendix \ref{sec:appendix_evaluation}, Table \ref{tab:testset_details}.
In our evaluation, we adopted a zero-shot approach, separately assessing knowledge-intensive tasks and Non-knowledge-intensive tasks,
and conduct simultaneous testing of Baichuan2-7B-Chat \cite{baichuan2023baichuan2} as a baseline.

\begin{table}[H]
\centering
\caption{Comparison of Baichuan2 and TongGu on a wider range of classical Chinese tasks. }
\resizebox{0.5\textwidth}{!}
{
\begin{tabular}{lcc}
\whline{1.1pt}
\textbf{Task} & \textbf{Baichuan2-7B-Chat} & \textbf{TongGu-7B-Instruct} \\
\hline
\rowcolor{gray!15}
\textit{Knowledge-intensive Tasks} & ACC ↑ & ACC ↑ \\
\hline
Source Retrieval & 0.00 & \textbf{96.67} \\
Author Retrieval & 3.33 & \textbf{100.00} \\
Previous Sentence Recitation & 0.00 & \textbf{46.67} \\
the Next Sentence Recitation & 10.00 & \textbf{83.33} \\
Entire Poem Recitation & 0.00 & \textbf{96.67} \\
\hline
\rowcolor{gray!15}
\textit{Non-knowledge-intensive Tasks} & PPL ↓ & PPL ↓ \\
\hline
Grammar & 5.63 & \textbf{4.84} \\
Ancient Poetry Writing & 10.44 & \textbf{8.25} \\
Couplet & \textbf{26.63} & 43.25 \\
Classical Chinese to Modern Chinese & 60.00 & \textbf{10.75} \\
Modern Chinese to Classical Chinese & 21.75 & \textbf{20.75} \\
Classical Chinese Writing & 20.75 & \textbf{14.00} \\
Poet Introduction & 17.75 & \textbf{11.06} \\
Analysis of Imagery & 17.13 & \textbf{3.40} \\
Knowledge of Sinology Q\&A & 9.06 & \textbf{8.75} \\
Idiom & 8.13 & \textbf{6.84} \\
Riddle & 13.38 & \textbf{12.75} \\
Xiehouyu & \textbf{10.75} & 27.50 \\
Flying Flower Order & \textbf{3.48} & 3.83 \\
Named Entity Recognition & 13.56 & \textbf{8.13} \\
Punctuation & 5.09 & \textbf{3.63} \\
Topic Classification & 9.81 & \textbf{7.16} \\
Word Explanation & 18.25 & \textbf{14.44} \\
Reading Comprehension & 2.22 & \textbf{1.88} \\
Poetry Appreciation & 12.56 & \textbf{10.75} \\
\whline{1.1pt}
\end{tabular}
}
\label{tab:result_all_tasks}
\end{table}
\vspace{-5pt}

\begin{table*}[ht]
\centering
\caption{Ablation study of various training strategies.}
\resizebox{0.8\textwidth}{!}{
\begin{tabular}{c|c|ccccc|c}
\whline{1.1pt}
\multirow{2}{*}{\textbf{Method}} & \multicolumn{1}{c|}{\textbf{Data-hungry Tasks}} & \multicolumn{5}{c|}{\textbf{Data-efficient Tasks}} & \multirow{2}{*}{\textbf{Avg. ↑}} \\
& Translation ↑ & Classifications ↑ & NER ↑ & Punctuation ↑ & Retrieval ↑ & Avg. ↑ \\
\hline
FT & 38.57 & 72.42 & 79.66 & 90.92 & 65.71 & \underline{77.18} & \underline{69.46} \\
LoRA & \underline{49.13} & 42.43 & 18.29 & 89.31 & 33.04 & 45.77 & 46.44 \\
RAT (Ours) & \textbf{54.43} & 72.47 & 73.46 & 89.97 & 77.30 & \textbf{78.30} & \textbf{73.53} \\
\whline{1.1pt}
\end{tabular}
}
\label{tab:ablation_rat}
\end{table*}

\begin{table*}[!t]
\centering
\caption{Ablation study on the impact of different \textit{N} used in RAT.}
\resizebox{0.8\textwidth}{!}{
\begin{tabular}{c|c|ccccc|c}
\whline{1.1pt}
\multirow{2}{*}{\textbf{N}} & \multicolumn{1}{c|}{\textbf{Data-hungry Tasks}} & \multicolumn{5}{c|}{\textbf{Data-efficient Tasks}} & \multirow{2}{*}{\textbf{Avg. ↑}} \\
& Translation ↑ & Classifications ↑ & NER ↑ & Punctuation ↑ & Retrieval ↑ & Avg. ↑ \\
\hline
4 & \textbf{54.47} & 66.24 & 69.02 & 89.91 & 79.78 & \underline{76.24} & \underline{71.88} \\
8 & \underline{54.43} & 72.47 & 73.46 & 89.97 & 77.30 & \textbf{78.30} & \textbf{73.53} \\
16 & 50.48 & 74.00 & 74.83 & 87.93 & 54.75 & 72.88 & 68.40 \\
\whline{1.1pt}
\end{tabular}
}
\label{tab:ablation_rat_N}
\end{table*}

\begin{table*}[!t]
\centering
\caption{Ablation study on the effect of CCU-RAG.}
\resizebox{0.9\textwidth}{!}{
\begin{tabular}{c|ccccc|c|c}
\whline{1.1pt}
\multirow{2}{*}{\textbf{Method}} & \multicolumn{5}{c|}{\textbf{Non-knowledge-intensive Tasks}} & \multicolumn{1}{c|}{\textbf{Knowledge-intensive Tasks}} & \multirow{2}{*}{\textbf{Avg. ↑}} \\
 & Translation ↑ & Classifications ↑ & NER ↑ & Punctuation ↑ & Avg. & Retrieval ↑ & \\
\hline
Ours & 54.43 & 72.47 & 73.46 & 89.97 & \textbf{72.58} & \textbf{77.30} & \textbf{73.53} \\
w/o RAG & 53.16 & 72.43 & 73.89 & 88.53 & 72.00 & 21.03 & 61.81 \\
\whline{1.1pt}
\end{tabular}
}
\label{tab:ablation_rag}
\end{table*}

For knowledge-intensive tasks, we utilize accuracy as the metric. 
For non-knowledge-intensive tasks, we use perplexity (PPL) as the metric, where the question and answer are concatenated and input into the model to compute the PPL. 
Table \ref{tab:result_all_tasks} presents the test results.
TongGu-7B-Instruct outperforms Baichuan-7B-Chat in 21 out of 24 tasks, which substantiates the effectiveness of our incremental pre-training and the two-stage fine-tuning approach (RAT).
More results can be found in Appendix \ref{sec:appendix_response}.

\subsection{Ablation Study}
\textbf{Fine-tuning methods.} 
We compared our proposed Redundancy-Aware Tuning (RAT) with two other fine-tuning methods: full-parameter fine-tuning (FT) and Low-Rank Adaptation (LoRA), using the C$^{3}$bench.
The results are summarized in Table \ref{tab:ablation_rat}. The results show that the vanilla FT method performs well in learning new tasks but suffers from catastrophic forgetting. The LoRA method mitigates catastrophic forgetting to some extent but struggles to adapt effectively to new tasks. In contrast, our proposed RAT method outperforms both FT and LoRA in terms of mitigating catastrophic forgetting and learning new tasks effectively. Ablation results on a wider range of tasks can be found in Appendix \ref{sec:appendix_ablation}, Table \ref{tab:ablation_rat_all_tasks}.

\textbf{RAT specifications.} 
We investigate the impact of different values of $N$ on the model performance used in the RAT method, as shown in Table \ref{tab:ablation_rat_N}. Setting $N$ to 8 provides the best performance, hence we adopt it as the default strategy.

\textbf{The effectiveness of CCU-RAG.} 
We evaluated the impact of our proposed CCU-RAG method by comparing TongGu with and without CCU-RAG on the C$^{3}$bench dataset, with results summarized in Table \ref{tab:result_all_tasks}.
The results reveal that the CCU-RAG method significantly improves TongGu's performance on the knowledge-intensive task of source retrieval, without diminishing the performance of Non-knowledge-intensive tasks such as punctuation and named entity recognition. Results on a wider range of tasks are given in Appendix \ref{sec:appendix_ablation}, Table \ref{tab:ablation_rag_all_tasks}.

\section{Conclusion}
In this paper, we introduce TongGu, a new state-of-the-art LLM specifically for Classical Chinese Understanding (CCU). 
Our contributions include the development of the ACCN-INS dataset, which serves as the first publicly accessible CCU instruction dataset, and the introduction of innovative techniques such as Redundancy-Aware Tuning (RAT) and CCU-RAG (Retrieval-Augmented Generation). Through extensive experiments and evaluations, we have demonstrated TongGu's superior performance in diverse CCU tasks, surpassing existing LLMs by a large margin in both knowledge-intensive and non-knowledge-intensive tasks.  We believe that TongGu and the AC-INS dataset will serve as valuable resources for future endeavors in the CCU research community.

\section{Limitations}
TongGu's performance heavily relies on the quality and quantity of the instruction-tuning dataset (ACCN-INS). 
The ACCN-INS dataset, while comprehensive, may not capture all variations in Classical Chinese texts. 
The RAT fine-tuning techniques, although effective, may still face challenges in mitigating catastrophic forgetting. 
Despite using CCU-RAG, the model may still produce hallucinations.
Overcoming these limitations and expanding research in CCU will drive progress in understanding classical languages and cultural heritage.

\section{Ethical and Social Implications}
TongGu has been trained to focus on processing ancient texts related to Chinese culture. 
Despite a series of cleaning and review processes applied to the training data, there may still be factual errors present, which could lead TongGu, like other large language models, to generate misleading information or harmful content that contains factual errors. 
In the future, we will continue to fine-tune and release updated versions as we progress in addressing these issues.

\section*{Acknowledgments}
This research is supported in part by National Natural Science Foundation of China (Grant No.: 62441604, 62476093).

\bibliography{custom}

\begin{thebibliography}{36}
\providecommand{\natexlab}[1]{#1}

\bibitem[{Asai et~al.(2024)Asai, Wu, Wang, Sil, and Hajishirzi}]{asai2024selfrag}
Akari Asai, Zeqiu Wu, Yizhong Wang, Avirup Sil, and Hannaneh Hajishirzi. 2024.
\newblock \href {https://openreview.net/forum?id=hSyW5go0v8} {Self-{RAG}: Learning to retrieve, generate, and critique through self-reflection}.
\newblock In \emph{Proc. ICLR}.

\bibitem[{Baichuan(2023)}]{baichuan2023baichuan2}
Baichuan. 2023.
\newblock \href {https://arxiv.org/abs/2309.10305} {Baichuan 2: Open large-scale language models}.
\newblock \emph{arXiv preprint arXiv:2309.10305}.

\bibitem[{Borgeaud et~al.(2022)Borgeaud, Mensch, Hoffmann, Cai, Rutherford, Millican, Van Den~Driessche, Lespiau, Damoc, Clark et~al.}]{borgeaud2022retro}
Sebastian Borgeaud, Arthur Mensch, Jordan Hoffmann, Trevor Cai, Eliza Rutherford, Katie Millican, George~Bm Van Den~Driessche, Jean-Baptiste Lespiau, Bogdan Damoc, Aidan Clark, et~al. 2022.
\newblock Improving language models by retrieving from trillions of tokens.
\newblock In \emph{Proc. ICML}, pages 2206--2240. PMLR.

\bibitem[{Brown et~al.(2020)Brown, Mann, Ryder, Subbiah, Kaplan, Dhariwal, Neelakantan, Shyam, Sastry, Askell et~al.}]{brown2020gpt3}
Tom~B Brown, Benjamin Mann, Nick Ryder, Melanie Subbiah, Jared Kaplan, Prafulla Dhariwal, Arvind Neelakantan, Pranav Shyam, Girish Sastry, Amanda Askell, et~al. 2020.
\newblock Language models are few-shot learners.
\newblock In \emph{Proc. NeurIPS}, pages 1877--1901.

\bibitem[{Cao et~al.(2024)Cao, Shi, Peng, Liu, and Jin}]{cao2024c3bench}
Jiahuan Cao, Yongxin Shi, Dezhi Peng, Yang Liu, and Lianwen Jin. 2024.
\newblock \href {https://arxiv.org/abs/2405.17732} {C$^{3}$bench: A comprehensive classical {C}hinese understanding benchmark for large language models}.
\newblock \emph{Preprint}, arXiv:2405.17732.

\bibitem[{Chang et~al.(2021)Chang, Shiue, Yeh, and Demberg}]{chang2021time-aware}
Ernie Chang, Yow-Ting Shiue, Hui-Syuan Yeh, and Vera Demberg. 2021.
\newblock {Time-Aware} ancient {C}hinese text translation and inference.
\newblock In \emph{Proc. LChange}, pages 1--6.

\bibitem[{Chang et~al.(2023)Chang, Dongbo, Zhixiao, Die, Mengcheng, Litao, Si, Bin, Jiangfeng, Hai et~al.}]{chang2023sikugpt}
Liu Chang, Wang Dongbo, Zhao Zhixiao, Hu~Die, Wu~Mengcheng, Lin Litao, Shen Si, Li~Bin, Liu Jiangfeng, Zhang Hai, et~al. 2023.
\newblock {SikuGPT}: A generative pre-trained model for intelligent information processing of ancient texts from the perspective of digital humanities.
\newblock \emph{arXiv preprint arXiv:2304.07778}.

\bibitem[{Chowdhery et~al.(2023)Chowdhery, Narang, Devlin, Bosma, Mishra, Roberts, Barham, Chung, Sutton, Gehrmann et~al.}]{chowdhery2023palm}
Aakanksha Chowdhery, Sharan Narang, Jacob Devlin, Maarten Bosma, Gaurav Mishra, Adam Roberts, Paul Barham, Hyung~Won Chung, Charles Sutton, Sebastian Gehrmann, et~al. 2023.
\newblock {PALM}: Scaling language modeling with pathways.
\newblock \emph{JMLR}, 24(240):1--113.

\bibitem[{Chung et~al.(2024)Chung, Hou, Longpre, Zoph, Tay, Fedus, Li, Wang, Dehghani, Brahma et~al.}]{chung2022flan-t5}
Hyung~Won Chung, Le~Hou, Shayne Longpre, Barret Zoph, Yi~Tay, William Fedus, Yunxuan Li, Xuezhi Wang, Mostafa Dehghani, Siddhartha Brahma, et~al. 2024.
\newblock Scaling instruction-finetuned language models.
\newblock \emph{JMLR}, 25(70):1--53.

\bibitem[{Gao et~al.(2023)Gao, Xiong, Gao, Jia, Pan, Bi, Dai, Sun, and Wang}]{gao2023rag-servey1}
Yunfan Gao, Yun Xiong, Xinyu Gao, Kangxiang Jia, Jinliu Pan, Yuxi Bi, Yi~Dai, Jiawei Sun, and Haofen Wang. 2023.
\newblock Retrieval-augmented generation for large language models: A survey.
\newblock \emph{arXiv preprint arXiv:2312.10997}.

\bibitem[{{Garychowcmu}(2019)}]{daizhige_github}
{Garychowcmu}. 2019.
\newblock Daizhigev20.
\newblock \url{https://github.com/garychowcmu/daizhigev20}.

\bibitem[{Gromov et~al.(2024)Gromov, Tirumala, Shapourian, Glorioso, and Roberts}]{gromov2024prune1}
Andrey Gromov, Kushal Tirumala, Hassan Shapourian, Paolo Glorioso, and Daniel~A Roberts. 2024.
\newblock The unreasonable ineffectiveness of the deeper layers.
\newblock \emph{arXiv preprint arXiv:2403.17887}.

\bibitem[{Guu et~al.(2020)Guu, Lee, Tung, Pasupat, and Chang}]{guu2020realm}
Kelvin Guu, Kenton Lee, Zora Tung, Panupong Pasupat, and Mingwei Chang. 2020.
\newblock Retrieval augmented language model pre-training.
\newblock In \emph{Proc. ICML}, pages 3929--3938. PMLR.

\bibitem[{Han et~al.(2018)Han, Xu, and Qiao}]{han2018chinese-term}
Xiaowei Han, Lizhen Xu, and Feng Qiao. 2018.
\newblock {CNN-BiLSTM-CRF} model for term extraction in {C}hinese corpus.
\newblock In \emph{Proc. WISA}, pages 267--274. Springer.

\bibitem[{Izacard et~al.(2023)Izacard, Lewis, Lomeli, Hosseini, Petroni, Schick, Dwivedi-Yu, Joulin, Riedel, and Grave}]{izacard2022atlas}
Gautier Izacard, Patrick Lewis, Maria Lomeli, Lucas Hosseini, Fabio Petroni, Timo Schick, Jane Dwivedi-Yu, Armand Joulin, Sebastian Riedel, and Edouard Grave. 2023.
\newblock Atlas: Few-shot learning with retrieval augmented language models.
\newblock \emph{Proc. JMLR}, 24(251):1--43.

\bibitem[{Jiang et~al.(2023)Jiang, Wang, Cao, Gao, and Jin}]{jiang2023C2C}
Zongyuan Jiang, Jiapeng Wang, Jiahuan Cao, Xue Gao, and Lianwen Jin. 2023.
\newblock Towards better translations from classical to modern {Chinese}: A new dataset and a new method.
\newblock In \emph{Proc. NLPCC}, pages 387--399. Springer.

\bibitem[{Lewis et~al.(2020)Lewis, Perez, Piktus, Petroni, Karpukhin, Goyal, K{\"u}ttler, Lewis, Yih, Rockt{\"a}schel et~al.}]{lewis2020rag}
Patrick Lewis, Ethan Perez, Aleksandra Piktus, Fabio Petroni, Vladimir Karpukhin, Naman Goyal, Heinrich K{\"u}ttler, Mike Lewis, Wen-tau Yih, Tim Rockt{\"a}schel, et~al. 2020.
\newblock Retrieval-augmented generation for knowledge-intensive {NLP} tasks.
\newblock \emph{Proc. NeurIPS}, 33:9459--9474.

\bibitem[{Li and Sun(2009)}]{li2009punctuation}
Zhongguo Li and Maosong Sun. 2009.
\newblock Punctuation as implicit annotations for chinese word segmentation.
\newblock \emph{Computational Linguistics}, 35(4):505--512.

\bibitem[{{MOP-LIWU Community} and {MNBVC Team}(2023)}]{mnbvc}
{MOP-LIWU Community} and {MNBVC Team}. 2023.
\newblock Mnbvc: Massive never-ending bt vast chinese corpus.
\newblock \url{https://github.com/esbatmop/MNBVC}.

\bibitem[{OpenAI(2023)}]{openai2023gpt4}
OpenAI. 2023.
\newblock {GPT-4} technical report.
\newblock \emph{arXiv preprint arXiv:2303.08774}.

\bibitem[{Ouyang et~al.(2022)Ouyang, Wu, Jiang, Almeida, Wainwright, Mishkin, Zhang, Agarwal, Slama, Gray et~al.}]{ouyang2022instruct-GPT}
Long Ouyang, Jeffrey Wu, Xu~Jiang, Diogo Almeida, Carroll Wainwright, Pamela Mishkin, Chong Zhang, Sandhini Agarwal, Katarina Slama, Alex Gray, et~al. 2022.
\newblock Training language models to follow instructions with human feedback.
\newblock In \emph{Proc. NeurIPS}.

\bibitem[{Radford et~al.(2018)Radford, Narasimhan, Salimans, Sutskever et~al.}]{radford2018gpt-1}
Alec Radford, Karthik Narasimhan, Tim Salimans, Ilya Sutskever, et~al. 2018.
\newblock Improving language understanding by generative pre-training.

\bibitem[{Raffel et~al.(2020)Raffel, Shazeer, Roberts, Lee, Narang, Matena, Zhou, Li, and Liu}]{raffel2020T5}
Colin Raffel, Noam Shazeer, Adam Roberts, Katherine Lee, Sharan Narang, Michael Matena, Yanqi Zhou, Wei Li, and Peter~J Liu. 2020.
\newblock Exploring the limits of transfer learning with a unified text-to-text transformer.
\newblock \emph{JMLR}, 21(140):1--67.

\bibitem[{Roziere et~al.(2023)Roziere, Gehring, Gloeckle, Sootla, Gat, Tan, Adi, Liu, Remez, Rapin et~al.}]{roziere2023codellama}
Baptiste Roziere, Jonas Gehring, Fabian Gloeckle, Sten Sootla, Itai Gat, Xiaoqing~Ellen Tan, Yossi Adi, Jingyu Liu, Tal Remez, J{\'e}r{\'e}my Rapin, et~al. 2023.
\newblock Code {LLaMA}: Open foundation models for code.
\newblock \emph{arXiv preprint arXiv:2308.12950}.

\bibitem[{shareAI(2023)}]{sharegpt_hf}
shareAI. 2023.
\newblock {ShareGPT}-{C}hinese-{E}nglish-90k bilingual human-machine qa dataset.
\newblock \url{https://huggingface.co/datasets/shareAI/ShareGPT-Chinese-English-90k}.

\bibitem[{{Together Computer}(2023)}]{together2023redpajama}
{Together Computer}. 2023.
\newblock \href {https://github.com/togethercomputer/RedPajama-Data} {Redpajama: An open source recipe to reproduce llama training dataset}.

\bibitem[{Touvron et~al.(2023{\natexlab{a}})Touvron, Lavril, Izacard, Martinet, Lachaux, Lacroix, Rozi{\`e}re, Goyal, Hambro, Azhar et~al.}]{touvron2023llama}
Hugo Touvron, Thibaut Lavril, Gautier Izacard, Xavier Martinet, Marie-Anne Lachaux, Timoth{\'e}e Lacroix, Baptiste Rozi{\`e}re, Naman Goyal, Eric Hambro, Faisal Azhar, et~al. 2023{\natexlab{a}}.
\newblock {LLaMA}: Open and efficient foundation language models.
\newblock \emph{arXiv preprint arXiv:2302.13971}.

\bibitem[{Touvron et~al.(2023{\natexlab{b}})Touvron, Martin, Stone, Albert, Almahairi, Babaei, Bashlykov, Batra, Bhargava, Bhosale et~al.}]{touvron2023llama2}
Hugo Touvron, Louis Martin, Kevin Stone, Peter Albert, Amjad Almahairi, Yasmine Babaei, Nikolay Bashlykov, Soumya Batra, Prajjwal Bhargava, Shruti Bhosale, et~al. 2023{\natexlab{b}}.
\newblock {LLaMA} 2: Open foundation and fine-tuned chat models.
\newblock \emph{arXiv preprint arXiv:2307.09288}.

\bibitem[{Wang et~al.(2023)Wang, Liu, Zhao, Shen, Liu, Li, Hu, Wu, Lin, Zhao et~al.}]{wang2023gujibert}
Dongbo Wang, Chang Liu, Zhixiao Zhao, Si~Shen, Liu Liu, Bin Li, Haotian Hu, Mengcheng Wu, Litao Lin, Xue Zhao, et~al. 2023.
\newblock {GujiBERT} and {GujiGPT}: Construction of intelligent information processing foundation language models for ancient texts.
\newblock \emph{arXiv preprint arXiv:2307.05354}.

\bibitem[{{Wptoux}(2023)}]{chunhua_hf}
{Wptoux}. 2023.
\newblock {Bloom-7B-chunhua}.
\newblock \url{https://huggingface.co/wptoux/bloom-7b-chunhua}.

\bibitem[{Wu et~al.(2023)Wu, Zhang, Zhang, Wang, and Xie}]{wu2023pmcllama}
Chaoyi Wu, Xiaoman Zhang, Ya~Zhang, Yanfeng Wang, and Weidi Xie. 2023.
\newblock {PMC-LLaMA}: Further finetuning {LLaMA} on medical papers.
\newblock \emph{arXiv preprint arXiv:2304.14454}.

\bibitem[{{Xunzi-LLM-of-Chinese-classics}(2024)}]{xunzi_github}
{Xunzi-LLM-of-Chinese-classics}. 2024.
\newblock {{XunziALLM}}.
\newblock \url{https://github.com/Xunzi-LLM-of-Chinese-classics/XunziALLM}.

\bibitem[{Yu and Wang(2020)}]{yu2020bert-ner}
Peng Yu and Xin Wang. 2020.
\newblock {BERT}-based named entity recognition in {C}hinese {Twenty-Four} {H}istories.
\newblock In \emph{Proc. WISA}, pages 289--301. Springer.

\bibitem[{Yunxiang et~al.(2023)Yunxiang, Zihan, Kai, Ruilong, and You}]{yunxiang2023chatdoctor}
Li~Yunxiang, Li~Zihan, Zhang Kai, Dan Ruilong, and Zhang You. 2023.
\newblock Chatdoctor: A medical chat model fine-tuned on {LLaMA} model using medical domain knowledge.
\newblock \emph{arXiv preprint arXiv:2303.14070}.

\bibitem[{Zhang et~al.(2022)Zhang, Roller, Goyal, Artetxe, Chen, Chen, Dewan, Diab, Li, Lin et~al.}]{zhang2022opt}
Susan Zhang, Stephen Roller, Naman Goyal, Mikel Artetxe, Moya Chen, Shuohui Chen, Christopher Dewan, Mona Diab, Xian Li, Xi~Victoria Lin, et~al. 2022.
\newblock {OPT}: Open pre-trained transformer language models.
\newblock \emph{arXiv preprint arXiv:2205.01068}.

\bibitem[{Zhao et~al.(2024)Zhao, Zhang, Yu, Wang, Geng, Fu, Yang, Zhang, and Cui}]{zhao2024rag-servey2}
Penghao Zhao, Hailin Zhang, Qinhan Yu, Zhengren Wang, Yunteng Geng, Fangcheng Fu, Ling Yang, Wentao Zhang, and Bin Cui. 2024.
\newblock Retrieval-augmented generation for {AI}-generated content: A survey.
\newblock \emph{arXiv preprint arXiv:2402.19473}.

\end{thebibliography}

\clearpage
\appendix

\section{Dataset Details}
\label{sec:appendix_data}

\begin{figure*}[ht]
\centering
\includegraphics[width=\textwidth]{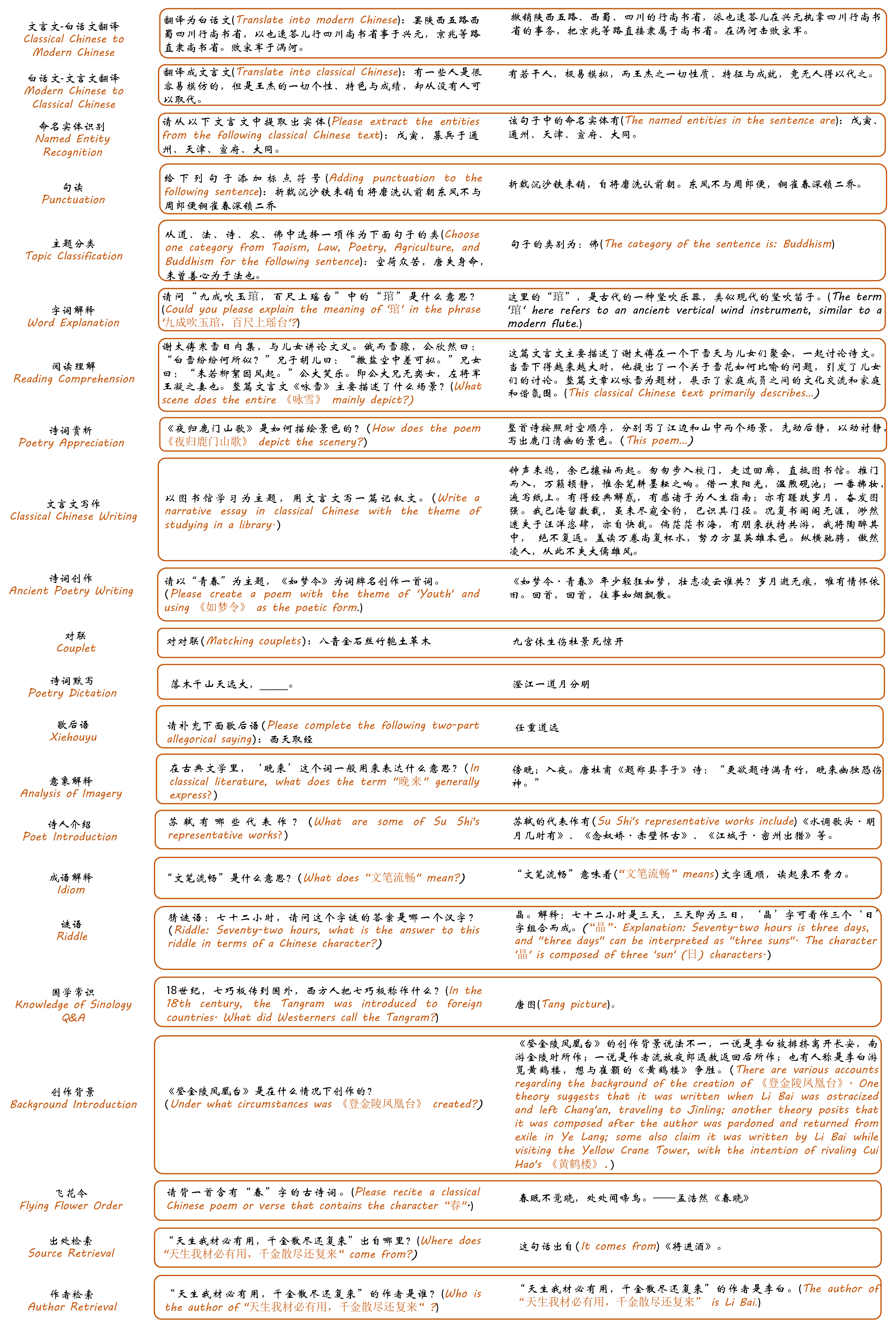}
\caption{Example of each task in ACCN-INS.}
\label{fig:task_example}
\end{figure*}

In this section, we present the details of our dataset. Figure \ref{fig:task_example} illustrates an example of each task in the dataset. The dimension and quantity of each task are shown in Table \ref{tab:task_details}.
It is notable that the ACCN-INS dataset will be released under the CC-BY-NC-SA-4.0 license, and the ShareGPT \cite{sharegpt_hf} dataset using in fine-tuning is under the apache-2.0 license.

\begin{table}[H]
\centering
\caption{Details of each task in ACCN-INS.}
\resizebox{0.5\textwidth}{!}
{
\begin{tabular}{lcc}
\whline{1.1pt}
\textbf{Task} & \textbf{Source} & \textbf{Quantity}\\
\hline
\rowcolor{gray!15}
\textit{comprehension Dimension} & & \\
\hline
Classical Chinese to Modern Chinese & Labeled data & 4,001,500 \\
Modern Chinese to Classical Chinese & Labeled data & 1,900 \\
Named Entity Recognition & Labeled data & 1,355 \\
Punctuation & Labeled data & 2,000 \\
Topic Classification & Labeled data & 1,200 \\
Word Explanation & Unlabeled data & 997 \\
Reading Comprehension & Unlabeled data & 363 \\
Appreciation & Unlabeled data & 600 \\
Grammar & Unlabeled data & 82 \\
\hline
\rowcolor{gray!15}
\textit{Generation Dimension} & & \\
\hline
Classical Chinese Writing & Unlabeled data & 165 \\
Ancient Poetry Writing & Unlabeled data & 165 \\
Couplet & Labeled data & 800 \\
\hline
\rowcolor{gray!15}
\textit{Knowledge Dimension} & & \\
\hline
Knowledge of Sinology Q\&A & Unlabeled data & 1,000 \\
Xiehouyu & Unlabeled data & 1,000 \\
Analysis of Imagery & Labeled data & 1,000 \\
Poet Introduction & Unlabeled data & 946 \\
Riddle & Unlabeled data & 463 \\
Idiom & Labeled data & 1,000 \\
Author Retrieval & Labeled data & 600 \\
Source Retrieval & Labeled data & 600 \\
Entire Poem Recitation & Labeled data & 600 \\
Previous Sentence Recitation & Labeled data & 600 \\
Next Sentence Recitation & Labeled data & 600 \\
Flying Flower Order & Labeled data & 56 \\
\hline
\textbf{total} & & 4,020,136 \\
\whline{1.1pt}
\end{tabular}
}
\label{tab:task_details}
\end{table}

\section{Training Details}
\label{sec:appendix_training}

In this section, we present the training details of TongGu. The hyper-parameter settings of incremental pre-training and instruction tuning are shown in \ref{tab:para_incremental_it}. 
All experiments are completed on 6 Nvidia A6000 GPUs.




\begin{table}[ht]
\centering
\caption{Hyper-parameter settings in incremental pre-training and instruction-tuning.}
\resizebox{0.8\linewidth}{!}{
\begin{tabular}{lcc}
\whline{1.1pt}
\multirow{3}{*}{\textbf{Hyper parameter}} & \multicolumn{2}{c}{\textbf{Value}} \\
\cline{2-3}
~ & \makecell[c]{Incremental\\Pretraining} & \makecell[c]{Instruction-\\tuning}\\
\hline
Precision & bf16 & bf16\\
Epoch & 1 & 3\\
Batch size & 576 & 192\\
Learning rate & 2e-5 & 2e-6\\
Weight decay & 0 & 0\\
Warmup ratio & 0 & 0\\
LR scheduler type & cosine & cosine\\
Optimizer & AdamW & AdamW\\
$\beta_1$ & 0.9 & 0.9\\
$\beta_2$ & 0.999 & 0.999\\
Max length & 2048 & 2048\\
\whline{1.1pt}
\end{tabular}}
\label{tab:para_incremental_it}
\end{table}


\section{Evaluation Details}
\label{sec:appendix_evaluation}

In this section, we detail the evaluation of TongGu on the C$^{3}$bench and our test dataset. The results, including the number of tasks and the evaluation metrics, are summarized in Tables \ref{tab:c3_details} and \ref{tab:testset_details}.

\begin{table}[H]
    \centering
    \caption{Details of C$^{3}$bench.}
    \resizebox{0.7\linewidth}{!}{
    \begin{tabular}{lcc}
    \whline{1.1pt}
    \textbf{Task} & \textbf{Metric} & \textbf{Quantity} \\
    \hline
    Classification & Accuracy & 10,000 \\
    Retrieval & Accuracy & 10,000 \\
    NER & F1-score & 10,000 \\
    Punctuation & F1-score & 10,000 \\
    Translation & BLEU & 10,000 \\
    \hline
    \textbf{Total} & - & 50,000 \\
    \whline{1.1pt}
    \end{tabular}}
    \label{tab:c3_details}
\end{table}

\begin{table}[H]
    \centering
    \caption{Details of our test dataset.}
    \resizebox{0.5\textwidth}{!}
    {
    \begin{tabular}{lcc}
    \whline{1.1pt}
    \textbf{Task} & \textbf{Metric} & \textbf{Quantity} \\
    \hline
    Source Retrieval & Accuracy & 30 \\
    Author Retrieval & Accuracy & 30\\
    Previous Sentence Recitation & Accuracy & 30 \\
    Next Sentence Recitation & Accuracy & 30 \\
    Entire Poem Recitation & Accuracy & 30 \\
    Grammar & PPL & 30 \\
    Ancient Poetry Writing & PPL & 30 \\
    Couplet & PPL & 100 \\
    Classical Chinese Writing & PPL & 30 \\
    Classical Chinese to Modern Chinese & PPL & 100 \\
    Modern Chinese to Classical Chinese & PPL & 100 \\
    Poet Introduction & PPL & 50 \\
    Analysis of Imagery & PPL & 100 \\
    Knowledge of Sinology Q\&A & PPL & 100 \\
    Idiom & PPL & 100 \\
    Riddle & PPL & 50 \\
    Xiehouyu & PPL & 100 \\
    Flying Flower Order & PPL & 30 \\
    Named Entity Recognition & PPL & 100 \\
    Punctuation & PPL & 100 \\
    Topic Classification & PPL & 100 \\
    Word Explanation & PPL & 100 \\
    Reading Comprehension & PPL & 30 \\
    Poetry Appreciation & PPL & 100 \\
    \hline
    \textbf{Total} & - & 1,600 \\
    \whline{1.1pt}
    \end{tabular}
    }
    \label{tab:testset_details}
\end{table}

\section{Sample Responses from Different Models}
\label{sec:appendix_response}

\begin{figure*}[ht]
\centering
\includegraphics[scale=0.5]{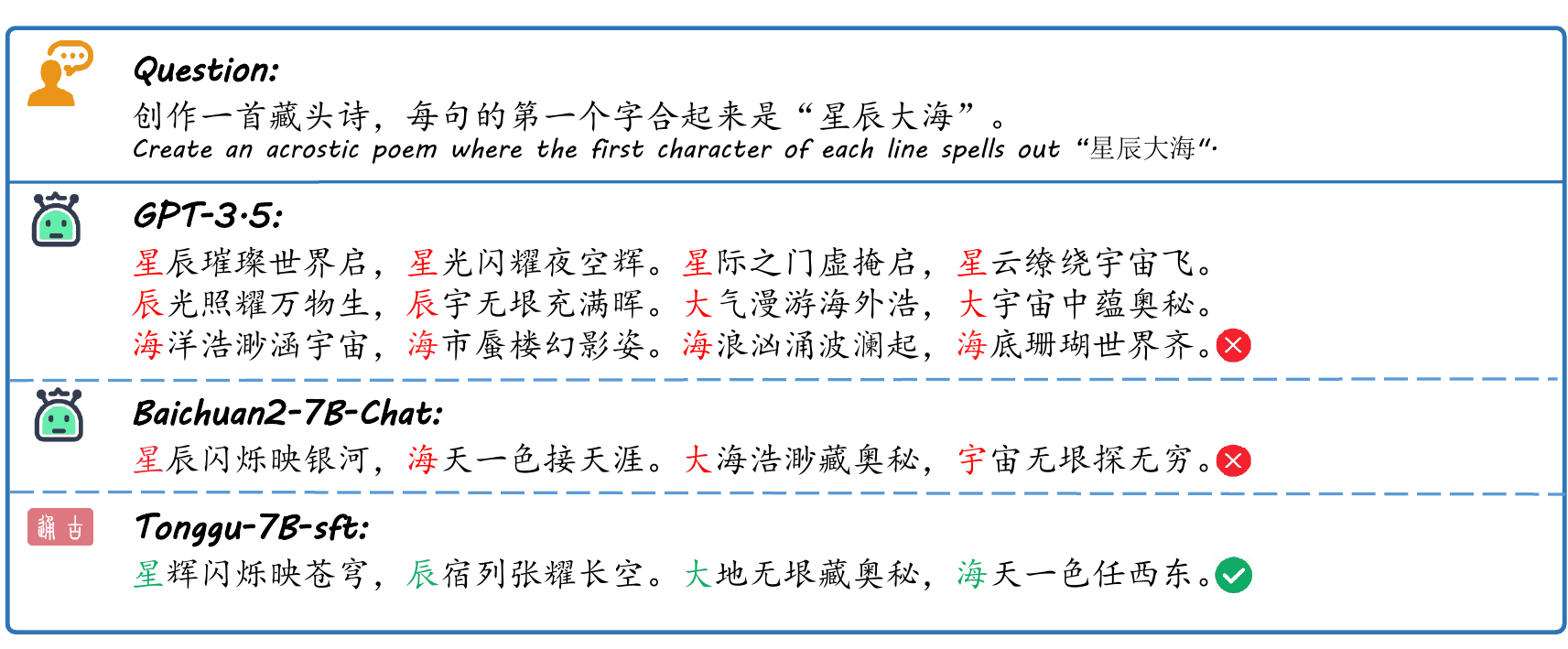}
\caption{Examples of responses to generation dimension tasks from different models.}
\label{fig:response_example_generation}
\end{figure*}

\begin{figure*}[ht]
\centering
\includegraphics[scale=0.5]{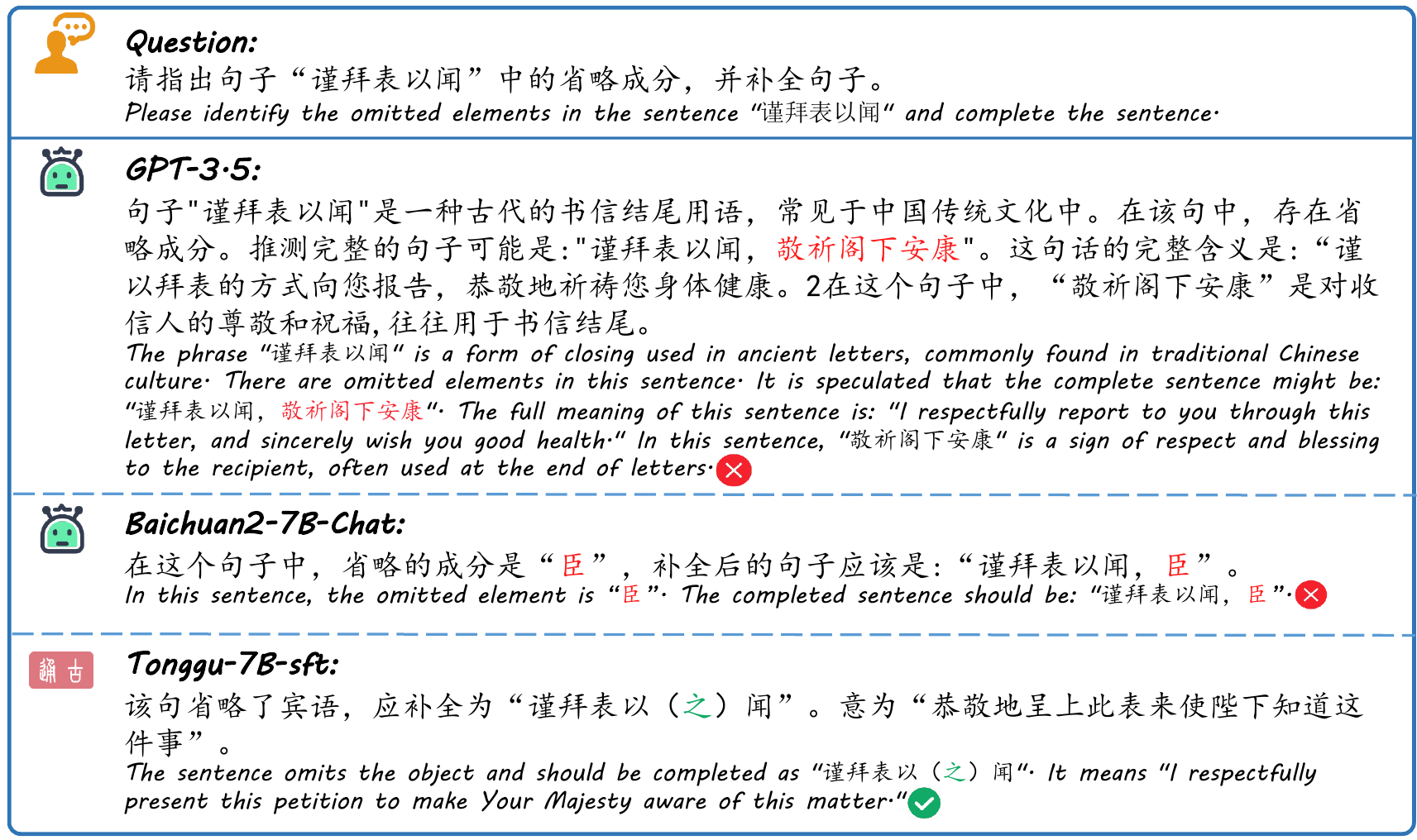}
\caption{Examples of responses to understanding dimension tasks from different models.}
\label{fig:response_example_understanding}
\end{figure*}

\begin{figure*}[ht]
\centering
\includegraphics[scale=0.5]{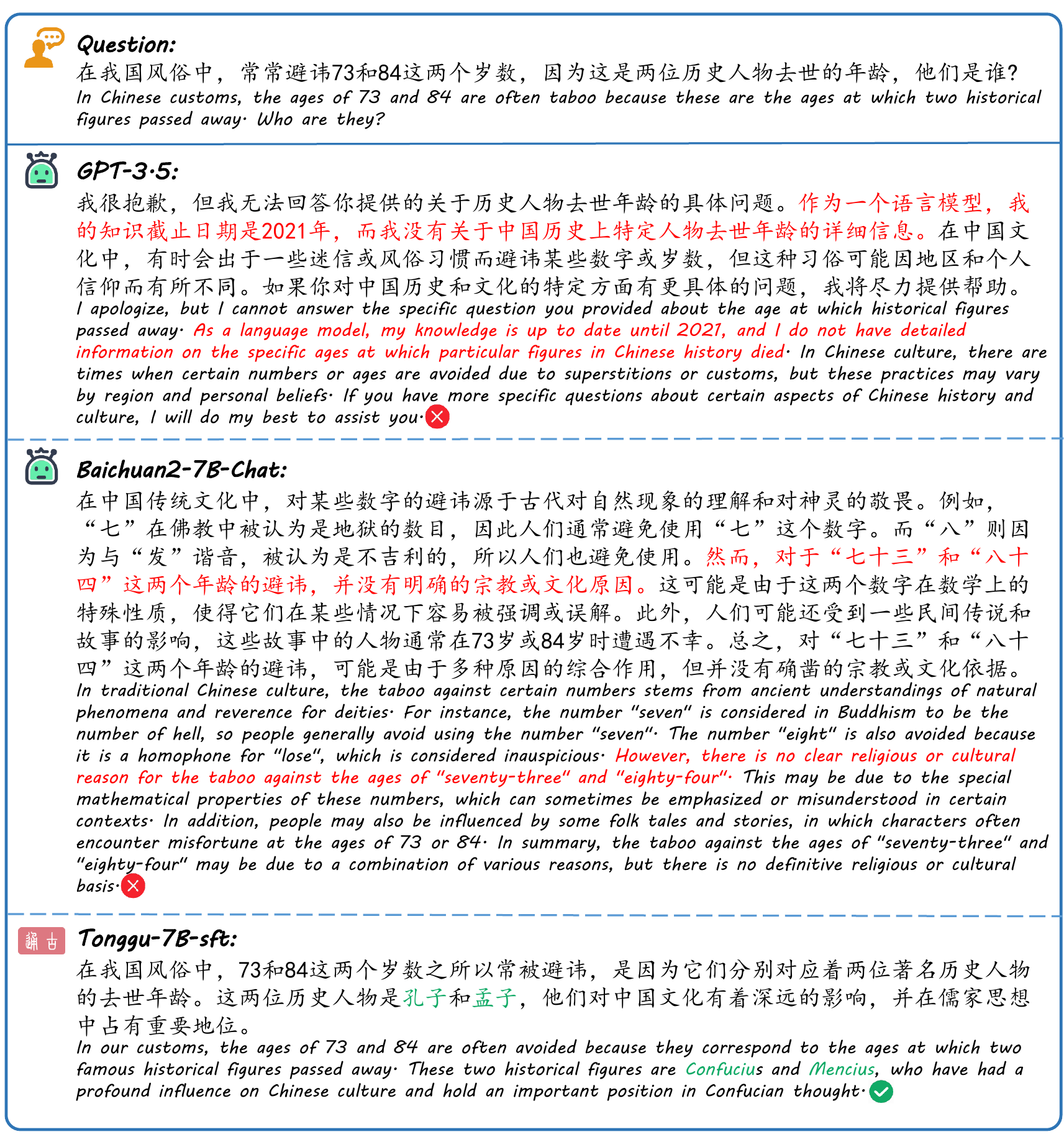}
\caption{Examples of responses to knowledge dimension tasks from different models.}
\label{fig:response_example_knowledge}
\end{figure*}

For each dimension, we present a task example, including the user's query, as well as the responses from Baihuan2-7B-Chat, GPT-3.5, and our TongGu model. It can be seen from Figure \ref{fig:response_example_generation} to \ref{fig:response_example_knowledge} that our TongGu model demonstrates excellent performance across the three dimensions of understanding, generation, and knowledge, proving the effectiveness of our data and training methods.

\section{Ablation Study Details}
\label{sec:appendix_ablation}

In this section, we present the ablation study in our test dataset. Table \ref{tab:ablation_rat_all_tasks} shows the performance of our RAT method and other training strategies, from which we can see that our RAT method not only alleviates catastrophic forgetting in data-hungry tasks but also generalizes well to data-efficient tasks. Table \ref{tab:ablation_rag_all_tasks} shown the ablation study on the effect of CCU-RAG, from which we can summarize that CCU-RAR significantly enhances the performance of knowledge-intensive tasks without compromising the Non-knowledge-intensive tasks.

\begin{table}[t]
\centering
\caption{Performance comparison of various training strategies on a wider range of classical Chinese tasks. Cells in yellow represent the retrieval tasks.}
\resizebox{0.5\textwidth}{!}
{
\begin{tabular}{lccc}
\whline{1.1pt}
\textbf{Task} & \textbf{FT} & \textbf{LoRA} & \textbf{RAT} \\
\hline
\rowcolor{gray!15}
\textit{Data-hungry Tasks} & & & \\
\hline
Classical Chinese to Modern Chinese & 12.75 & \underline{11.06} & \textbf{10.75} \\
\hline
\rowcolor{gray!15}
\textit{Data-Efficient Tasks} & & & \\
\hline
\rowcolor{yellow!15}
Source Retrieval & \textbf{96.67} & 0.00 &\textbf{96.67} \\
\rowcolor{yellow!15}
Author Retrieval & \underline{53.33} & 6.67 & \textbf{100.00} \\
\rowcolor{yellow!15}
Previous Sentence Recitation & \textbf{60.00} & 33.33 & \underline{46.67} \\
\rowcolor{yellow!15}
Next Sentence Recitation & \textbf{83.33} & 36.67 & \textbf{83.33} \\
\rowcolor{yellow!15}
Entire Poem Recitation & \textbf{96.67} & 66.67 &\textbf{96.67} \\
Grammar & \underline{5.10} & 5.25 & \textbf{4.84} \\
Ancient Poetry Writing & \textbf{7.34} & 12.18 & \underline{8.25} \\
Couplet & \textbf{33.00} & 45.75 & \underline{43.25} \\
Modern Chinese to Classical Chinese & \underline{18.88} & \textbf{18.00} & 20.75 \\
Classical Chinese Writing & \textbf{13.75} & 18.25 & \underline{14.00} \\
Poet Introduction & \textbf{8.94} & 13.56 & \underline{11.06} \\
Analysis of Imagery & 4.44 & \underline{3.63} & \textbf{3.41} \\
Knowledge of Sinology Q\&A & \underline{9.31} & 9.50 &\textbf{8.75} \\
Idiom & \textbf{6.53} & 7.09 & \underline{6.84} \\
Riddle & 13.81 & \underline{13.38} & \textbf{12.75} \\
Xiehouyu & \textbf{18.88} & \underline{24.25} & 27.50 \\
Flying Flower Order & \textbf{3.30} & 4.22 & \underline{3.83} \\
Named Entity Recognition & \underline{7.87} & \textbf{7.28} & 8.13 \\
Punctuation & 3.65 & \textbf{3.63} & \textbf{3.63} \\
Topic Classification & \underline{7.34} & 8.41 & \textbf{7.16} \\
Word Explanation & 15.63 & \textbf{13.38} & \underline{14.44} \\
Reading Comprehension & 2.11 & \underline{1.89} & \textbf{1.88} \\
Poetry Appreciation & 13.19 & \underline{12.93} & \textbf{12.75} \\
\whline{1.1pt}
\end{tabular}
}
\label{tab:ablation_rat_all_tasks}
\end{table}

\begin{table}[t]
\centering
\caption{Ablation study on the effect of CCU-RAG on a wider range of classical Chinese tasks.}
\resizebox{0.5\textwidth}{!}
{
\begin{tabular}{lcc}
\whline{1.1pt}
\textbf{Task} & \textbf{Ours} & \textbf{w/o RAG}  \\
\hline
\rowcolor{gray!15}
\textit{Knowledge-intensive Tasks} & ACC ↑ & ACC ↑ \\
\hline
Source Retrieval & \textbf{96.67} & 3.33 \\
Author Retrieval & \textbf{100.00} & 30.00\\
Previous Sentence Recitation & \textbf{46.67} & 0.00\\
Next Sentence Recitation & \textbf{83.33} & 33.33 \\
Entire Poem Recitation &\textbf{96.67} & 0.00 \\
\hline
\rowcolor{gray!15}
\textit{Non-knowledge-intensive Tasks} & PPL ↓ & PPL ↓ \\
\hline
Grammar & \textbf{4.84} & 4.89 \\
Ancient Poetry Writing & \textbf{8.25} & \textbf{8.25}\\
Couplet & 43.25 &\textbf{42.50} \\
Classical Chinese to Modern Chinese & \textbf{10.75} & \textbf{10.75} \\
Modern Chinese to Classical Chinese & \textbf{20.75} & 21.75\\
Classical Chinese Writing & \textbf{14.00} & 14.38 \\
Poet Introduction & 11.06 & \textbf{10.44} \\
Analysis of Imagery & \textbf{3.41} & 3.44\\
Knowledge of Sinology Q\&A & \textbf{8.75} & \textbf{8.75} \\
Idiom & \textbf{6.84} & 6.98 \\
Riddle & \textbf{12.75} & \textbf{12.75} \\
Xiehouyu & 27.50 & \textbf{25.75}\\
Flying Flower Order & \textbf{3.83} & 3.85 \\
Named Entity Recognition & 8.13 & \textbf{7.88}\\
Punctuation & 3.63 & \textbf{3.52} \\
Topic Classification & 7.16 & \textbf{6.78} \\
Word Explanation & 14.44 & \textbf{14.00} \\
Reading Comprehension & 1.88 & \textbf{1.52} \\
Poetry Appreciation & \textbf{12.75} & \textbf{12.75} \\
\whline{1.1pt}
\end{tabular}
}
\label{tab:ablation_rag_all_tasks}
\end{table}
\hfill

\end{document}